\documentclass{article}

\usepackage{mathtools}
\usepackage{amsmath,amsfonts,bm}

\def\1{\bm{1}}

\def\gD{{\mathcal{D}}}

\def\gI{{\mathcal{I}}}
\def\gJ{{\mathcal{J}}}

\def\gU{{\mathcal{U}}}

\def\gX{{\mathcal{X}}}
\def\gY{{\mathcal{Y}}}
\def\gZ{{\mathcal{Z}}}

\def\sP{{\mathbb{P}}}

\def\sR{{\mathbb{R}}}

\newcommand{\E}{\mathbb{E}}
\newcommand{\Ls}{\mathcal{L}}

\DeclareMathOperator*{\argmax}{arg\,max}
\DeclareMathOperator*{\argmin}{arg\,min}
\DeclareMathOperator*{\subjectto}{subject\;to}
\DeclareMathOperator*{\st}{s.t.}

\newcommand{\ftrans}{\ensuremath{f^{\mathrm{trans}}}}
\newcommand{\fodec}{\ensuremath{f^{\mathrm{odec}}}}
\newcommand{\frew}{\ensuremath{f^{\mathrm{rew}}}}
\newcommand{\fdec}{\ensuremath{f^{\mathrm{dec}}}}

\newcommand{\xinit}{x_{\rm init}}

\newcommand{\defeq}{\vcentcolon=}

\usepackage[T1]{fontenc}
\usepackage{graphicx}
\graphicspath{{./images/}}

\usepackage{xcolor}
\definecolor{mycolor}{RGB}{0,128,255}

\usepackage{wrapfig}
\usepackage{nicefrac}
\usepackage{microtype}
\usepackage{graphicx}
\usepackage{subfig}
\usepackage{booktabs}
\usepackage{tikz}

\usepackage[colorlinks=true,allcolors=mycolor,pageanchor=true,plainpages=false,pdfpagelabels,bookmarks,bookmarksnumbered]{hyperref}

\usepackage[nameinlink]{cleveref}
\Crefname{section}{Sect.}{Sects.}
\Crefname{appendix}{App.}{Apps.}

\usepackage{amssymb}
\DeclareSymbolFont{bbold}{U}{bbold}{m}{n}
\DeclareSymbolFontAlphabet{\mathbbold}{bbold}

\usepackage{amsthm}

\newtheorem{proposition}{Proposition}
\newtheorem{corollary}{Corollary}
\newtheorem{lemma}{Lemma}
\Crefname{algorithm}{Alg.}{Algs.}
\Crefname{proposition}{Prop.}{Props.}
\Crefname{lemma}{lem.}{Lems.}

\usepackage{algorithm}
\usepackage{algorithmicx}
\usepackage{algpseudocode}
\algnewcommand{\LeftCommentX}[1]{\Statex \(\triangleright\) #1}
\algnewcommand{\LeftComment}[1]{\State \(\triangleright\) #1}
\makeatother

\newcommand{\cblock}[3]{
  \hspace{-1.5mm}
  \begin{tikzpicture}
    [
    node/.style={square, minimum size=10mm, thick, line width=0pt},
    ]
    \node[fill={rgb,255:red,#1;green,#2;blue,#3}] () [] {};
  \end{tikzpicture}%
}

\usepackage{todonotes}

\usepackage{xspace}
\newcommand{\eg}{{\it e.g.}\xspace}
\newcommand{\ie}{{\it i.e.}\xspace}

\newcommand{\LML}{\ensuremath{\mathcal{L}}}

\usepackage[accepted]{icml2020}
\icmltitlerunning{The Differentiable Cross-Entropy Method}

\begin{document}

\twocolumn[
\icmltitle{The Differentiable Cross-Entropy Method}

\begin{icmlauthorlist}
\icmlauthor{Brandon Amos}{fair}
\icmlauthor{Denis Yarats}{fair,nyu}
\end{icmlauthorlist}

\icmlcorrespondingauthor{Brandon Amos}{bda@fb.com}
\icmlaffiliation{fair}{Facebook AI Research}
\icmlaffiliation{nyu}{New York University}

\icmlkeywords{Machine Learning, Reinforcement Learning, Continuous Control}

\vskip 0.3in
]

\printAffiliationsAndNotice{}

\begin{abstract}
  We study the cross-entropy method (CEM) for the non-convex
  optimization of a continuous and parameterized objective function
  and introduce a differentiable variant that
  enables us to differentiate the output of CEM with respect
  to the objective function's parameters.
  In the machine learning setting this brings CEM
  inside of the end-to-end learning pipeline where
  this has otherwise been impossible.
  We show applications in a synthetic energy-based structured
  prediction task and in non-convex continuous control.
  In the control setting we show how to embed optimal
  action sequences into a lower-dimensional space.
  DCEM enables us to fine-tune CEM-based controllers
  with policy optimization.
\end{abstract}

\section{Introduction}
Recent work in the machine learning community has shown how
optimization procedures can create new building-blocks
for the end-to-end machine learning pipeline
\citep{gould2016differentiating,johnson2016composing,amos2017input,amos2017optnet,domke2012generic,metz2016unrolled,finn2017model,zhang2019deep,belanger2017end,rusu2018meta,srinivas2018universal,amos2018differentiable,agrawal2019differentiable}.
In this paper we focus on the setting of optimizing
an \emph{unconstrained}, \emph{non-convex}, and \emph{continuous}
objective function
$f_\theta(x):\sR^n \times\Theta\rightarrow \sR$
as
\begin{equation}
\hat x \defeq \argmin_x f_\theta(x),
\label{eq:intro_argmin}
\end{equation}
where we assume $\hat x$ is unique and that
$f$ is parameterized by $\theta\in\Theta$
and has inputs $x\in\sR^n$.
\emph{If} it exists, \emph{some} \mbox{(sub-)derivative}
$\nabla_\theta \hat x$
is useful in the machine learning setting to make
the output of the optimization procedure end-to-end learnable.
For example, $\theta$ could parameterize a predictive model
that generates outcomes conditional on $x$.
End-to-end learning in these settings can be done by
defining a loss function $\Ls$ on top of $\hat x$
and taking gradient steps
$\nabla_\theta \Ls$.
If $f_\theta$ were \emph{convex} this gradient is easy to analyze
and compute when it exists and is unique
\citep{gould2016differentiating,johnson2016composing,amos2017input,amos2017optnet}.
Analyzing and computing a ``derivative'' through the
\emph{non-convex} $\argmin$ in
\cref{eq:intro_argmin}
is not as easy and is challenging in theory and practice.
The derivative may not exist or may be \emph{uninformative} in theory,
it might not be unique, and even if it does, the numerical solver being used
to compute the solution may not find a global or even local optimum of $f$.
One promising direction to sidestep these issues is to approximate
the $\argmin$ operation with an explicit optimization procedure
that is interpreted as another compute graph and \emph{unrolled} through,
\ie seen as a sequence of differentiable computations.
This is most commonly done with gradient descent as in
\citet{domke2012generic,metz2016unrolled,finn2017model,belanger2017end,rusu2018meta,srinivas2018universal,foerster2018learning,zhang2019deep}.
This approximation adds definition and structure to an
otherwise ill-defined desiderata at the cost of
biasing the gradients and enabling the learning procedure to
over-fit to the hyper-parameters of the optimization algorithm,
such as the number of gradient steps or the learning rate.

In this paper we show how to use the \emph{cross-entropy method} (CEM)
\citep{rubinstein1997optimization,de2005tutorial}
to approximate the derivative through an
\emph{unconstrained}, \emph{non-convex}, and \emph{continuous}
$\argmin$.
CEM for optimization is a \emph{zeroth-order optimizer} and works by
generating a sequence of samples from the objective function.
We show a simple and computationally negligible
way of making CEM differentiable that we call DCEM
by using the smooth top-$k$ operation from \citet{amos2019limited}.
This also brings CEM into the end-to-end learning process in scenarios
such as control where there is otherwise a disconnection between the
objective that is being learned and the objective that is induced by
deploying CEM on top of those models.

We first study DCEM in a simple non-convex
\emph{energy-based learning} setting for regression.
We contrast using unrolled gradient descent and DCEM
for optimizing over a SPEN \citep{belanger2016structured}.
We show that unrolling through gradient descent in this
setting over-fits to the number of gradient steps taken
and that DCEM generates a more reasonable energy surface.

We next focus on using DCEM in
the context of \emph{non-convex continuous control}
as a \emph{differentiable policy class} that is end-to-end learnable.
This setting is especially interesting as vanilla CEM is
the state-of-the-art method for solving the
control optimization problem with \emph{neural network transition dynamics}
as in \citet{chua2018deep,hafner2018learning}.
We show that DCEM is useful for embedding \emph{action sequences}
into a lower-dimensional space to make solving the
control optimization process significantly less
computationally and memory expensive.
This controller induces a \emph{differentiable policy class}
parameterized by the model-based components.
DCEM is a solution to the
\emph{objective mismatch problem}
in model-based control \citep{lambert2020objective},
which is the issue that arises when training model-based components with
the objective of maximizing the data likelihood but then using the
model-based components for the objective of control.
We use PPO \citep{schulman2017proximal} to
\emph{fine-tune} the model-based components, demonstrating that it \emph{is}
possible to use standard policy learning for model-based RL components in
addition to maximum-likelihood fitting.

\section{Background and Related Work}
\subsection{Differentiable optimization-based modeling in machine learning}
\emph{Optimization-based modeling} is a way of integrating specialized
operations and domain knowledge into end-to-end machine learning pipelines,
typically in the form of a parameterized $\argmin$ operation.
\emph{Convex}, \emph{constrained}, and \emph{continuous} optimization problems,
\eg as in
\citet{gould2016differentiating,johnson2016composing,amos2017input,amos2017optnet,agrawal2019differentiable},
capture many standard layers as special cases and
can be differentiated through by applying the \emph{implicit function theorem}
to a set of optimality conditions from convex optimization theory,
such as the \emph{KKT conditions}.
\emph{Non-convex} and \emph{continuous} optimization problems,
\emph{e.g.}\ as in
\citet{domke2012generic,belanger2016structured,metz2016unrolled,finn2017model,belanger2017end,rusu2018meta,srinivas2018universal,foerster2018learning,amos2018differentiable,pedregosa2016hyperparameter,jenni2018deep,rajeswaran2019meta,zhang2019deep},
are more difficult to differentiate through.
Differentiation is typically done by \emph{unrolling gradient descent}
or applying the implicit function theorem to \emph{some}
set of optimality conditions, sometimes forming a locally
convex approximation to the larger non-convex problem.
\emph{Unrolling gradient descent} is the most common way and approximates
the $\argmin$ operation with gradient descent for the forward pass and
interprets the operations as another compute graph for the backward pass
that can all be differentiated through.
In contrast to these works, we show how \emph{continuous} and
\emph{non-convex} $\argmin$ operations can also be approximated with
the cross-entropy method.

\subsection{Embedding domains for optimization problems}
Oftentimes the \emph{solution space} of high-dimensional
optimization problems may have structural properties that
an optimizer can exploit to find a better solution or
to find the solution quicker than an otherwise na\"ive optimizer.
Meta-learning approaches such as LEO \citep{rusu2018meta} and
CAVIA \citep{zintgraf2019fast} turn the optimization
problem for adaptation in a high-dimensional parameter space into
a lower-dimensional latent embedded optimization problem.
In the context of Bayesian optimization this has been
explored with random feature embeddings, hand-coded embeddings,
and auto-encoder-learned embeddings
\citep{antonova2019bayesian,oh2018bock,calandra2016manifold,wang2016bayesian,garnett2013active,salem2019sequential,kirschner2019adaptive}.
\citet{luo2018neural} and \citet{gomez2018automatic} turn
discrete search problems for architecture search and molecular design,
respectively, into embedded continuous optimization problems.
We show that DCEM is another reasonable way of learning
an embedded domain for exploiting the structure in and efficiently
solving larger optimization problems, with the significant advantage
of DCEM being that the latent space is directly learned to be
optimized over as part of the end-to-end learning pipeline.

\subsection{RL and Control}
\label{ref:bg:rl}
High-dimensional non-convex optimization problems that have a lot of
structure in the solution space naturally arise in the control setting
where the controller seeks to optimize the same objective in
the same controller dynamical system from different starting states.
This has been investigated in, \eg,
planning
\citep{ichter2018learning,ichter2019robot,mukadam2018continuous,kurutach2018learning,srinivas2018universal,yu2019unsupervised,lynch2019learning},
and policy distillation \citep{wang2019exploring}.
\citet{chandak2019learning} shows how to learn an action space
for model-free learning and \citet{co2018self,antonova2019bayesian}
embed \emph{action sequences} with a VAE.
There has also been a lot of work on learning reasonable
latent \emph{state space} representations \citep{tasfi2018dynamic,zhang2018solar,gelada2019deepmdp,miladinovic2019disentangled}
that may have structure imposed to make it more controllable
\citep{watter2015embed,banijamali2017robust,ghosh2018learning,anand2019unsupervised,levine2019prediction,singh2019learning}.
In contrast to these works, we learn how to encode action sequences directly
with DCEM instead of auto-encoding the sequences.
This has the advantages of 1) never requiring the expensive expert's
solution to the control optimization problem, 2) potentially being able to surpass
the performance of an expert controller that uses the full action space, and
3) being end-to-end learnable through the controller for the
purpose of finding a latent space of sequences that DCEM
is good at searching over.

Another direction the RL and control communities has been pursuing is
on the combination of model-based and model-free methods by
differentiating through model-based components
\citet{bansal2017goal} does this with Bayesian optimization and
locally linear models.
\citet{okada2017path,pereira2018mpc} makes path integral control
\citep{theodorou2010generalized} differentiable.
\citet{agrawal2019learning} considers a class of convex controllers
and differentiates through them with \citet{agrawal2019differentiable}.
\citet{amos2018differentiable} proposes differentiable MPC
and only do imitation learning on the cartpole
and pendulum tasks with known or lightly-parameterized dynamics ---
in contrast, we are able to
1) scale our differentiable controller up to the cheetah
and walker tasks,
2) use neural network dynamics inside of our controller, and
3) backpropagate a policy loss through the output of
our controller and into the internal components.

\section{The Differentiable Cross-Entropy Method}
The cross-entropy method (CEM) \citep{rubinstein1997optimization,de2005tutorial}
is an algorithm to solve optimization problems
in the form of \cref{eq:intro_argmin}.
CEM is an \emph{iterative} and \emph{zeroth-order} solver
that uses a sequence of \emph{parametric sampling distributions} $g_\phi$
defined over the domain $\sR^n$, such as Gaussians.
Given a \emph{sampling distribution} $g_\phi$, the hyper-parameters of CEM are
the number of \emph{candidate points} sampled in each iteration $N$,
the number of \emph{elite candidates} $k$ to use to
fit the new sampling distribution to, and
the number of iterations $T$.
The iterates of CEM are the \emph{parameters} $\phi$ of the
sampling distribution.
CEM starts with an \emph{initial} sampling distribution
$g_{\phi_1}(X)\in\sR^n$,
and in each iteration $t$ generates $N$ samples from the domain
$\left[X_{t,i}\right]_{i=1}^N \sim g_{\phi_t}(\cdot)$,
evaluates the function at those points $v_{t,i}\defeq f_\theta(X_{t,i})$,
and re-fits the sampling distribution to the top-$k$ samples by
solving the maximum-likelihood problem\footnote{%
CEM's name comes from
\cref{eq:cem-update} more generally optimizing the
cross-entropy measure between two distributions.}
\begin{equation}
  \phi_{t+1} \defeq
    \argmax_\phi \sum_i \mathbbold{1}{\{v_{t,i} \leq \pi(v_t)_k\}} \log g_\phi(X_{t,i}),
   \label{eq:cem-update}
\end{equation}
where the indicator $\mathbbold{1}\{P\}$ is 1 if $P$ is true and 0 otherwise,
$g_\phi(X)$ is the likelihood of $X$ under the distribution
$g_\theta$,
and $\pi(x)$ sorts $x\in\sR^n$ in ascending order so that
$$\pi(x)_1 \leq \pi(x)_2 \leq \ldots \leq \pi(x)_n.$$
We can then map from the final distribution $g_{\phi_T}$ back
to the domain by taking the mean of it,
\ie $\hat x \defeq \E[g_{\phi_{T+1}}(\cdot)]$.
In some settings, the best sample can be returned as
$\hat x$.

\begin{proposition}
For multivariate isotropic Gaussian sampling distributions
we have that $\phi=\{\mu, \sigma^2\}$ and
\cref{eq:cem-update} has a closed-form solution
given by the sample mean and variance of the top-$k$ samples
as $\mu_{t+1}\defeq \nicefrac{1}{k} \sum_{i\in\gI_t} X_{t,i}$
and $\sigma^2_{t+1}\defeq \nicefrac{1}{k} \sum_{i\in\gI_t} \left(X_{t,i}-\mu_{t+1}\right)^2$,
where the top-$k$ indexing set is $\gI_t\defeq \{i: v_{t,i} \leq \pi(v_t)_k\}$.
\label{prop:vanilla-cem}
\end{proposition}

This is well-known and is discussed in,
\eg, \citet{friedman2001elements}.
We present this here to make the connections
between CEM and DCEM clearer.

Differentiating through CEM's output with respect to the objective
function's parameters with $\nabla_\theta \hat x$ is useful, \eg,
to bring CEM into the end-to-end learning process in cases
where there is otherwise a disconnection between the objective
that is being learned and the objective that is induced by
deploying CEM on top of those models.
In the vanilla form presented above the
top-$k$ operation in \cref{eq:cem-update} makes $\hat x$
non-differentiable with respect to $\theta$.
The function samples can usually be differentiated through
with some estimator \citep{mohamed2019monte} such
as the \emph{reparameterization trick}
\citep{kingma2013auto,rezende2014stochastic,titsias2014doubly},
which we use in all of our experiments.

\begin{figure}[t]
  \centering
  \includegraphics[width=.24\textwidth]{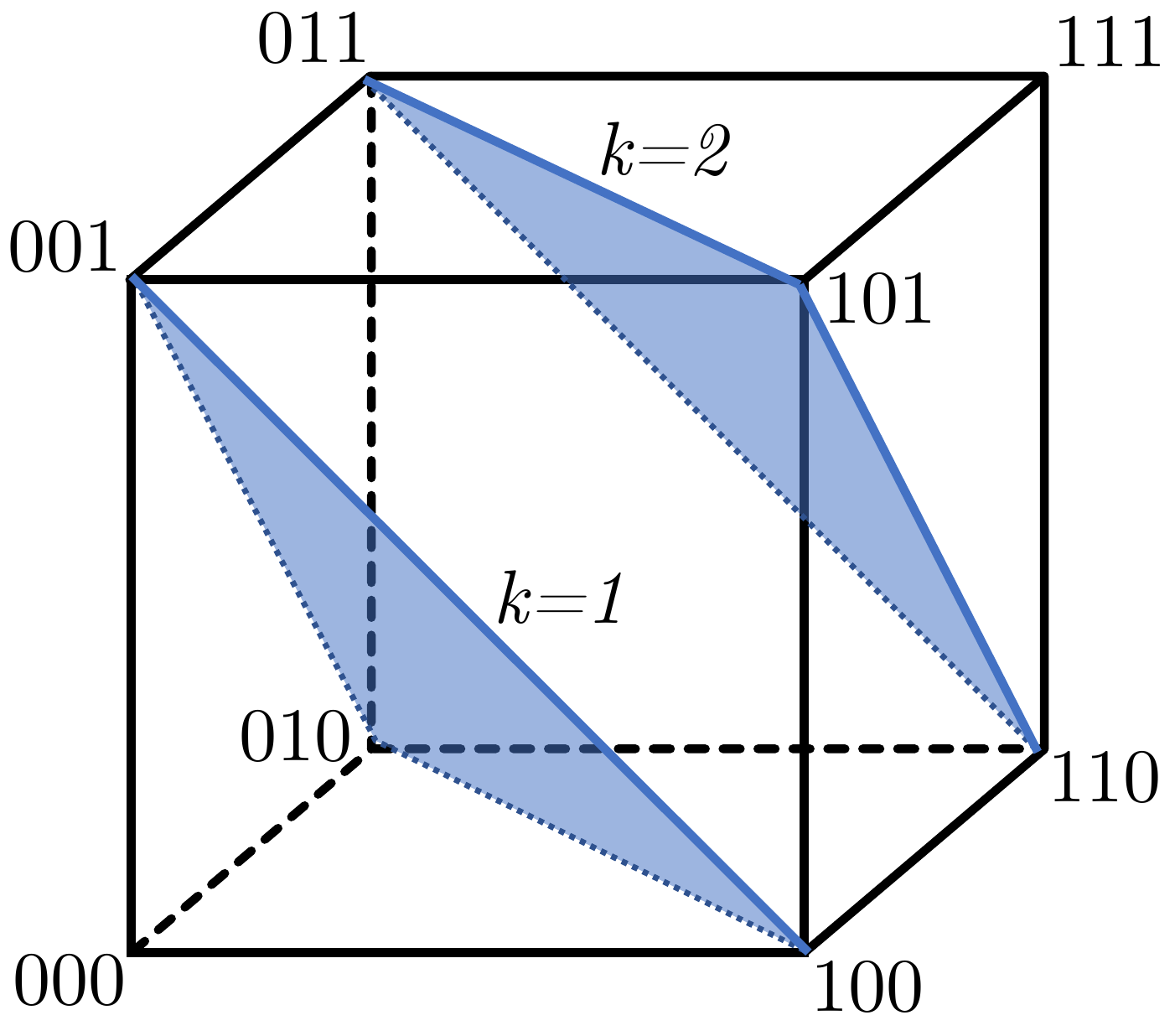}
  \hspace{-2mm}
  \includegraphics[width=.24\textwidth]{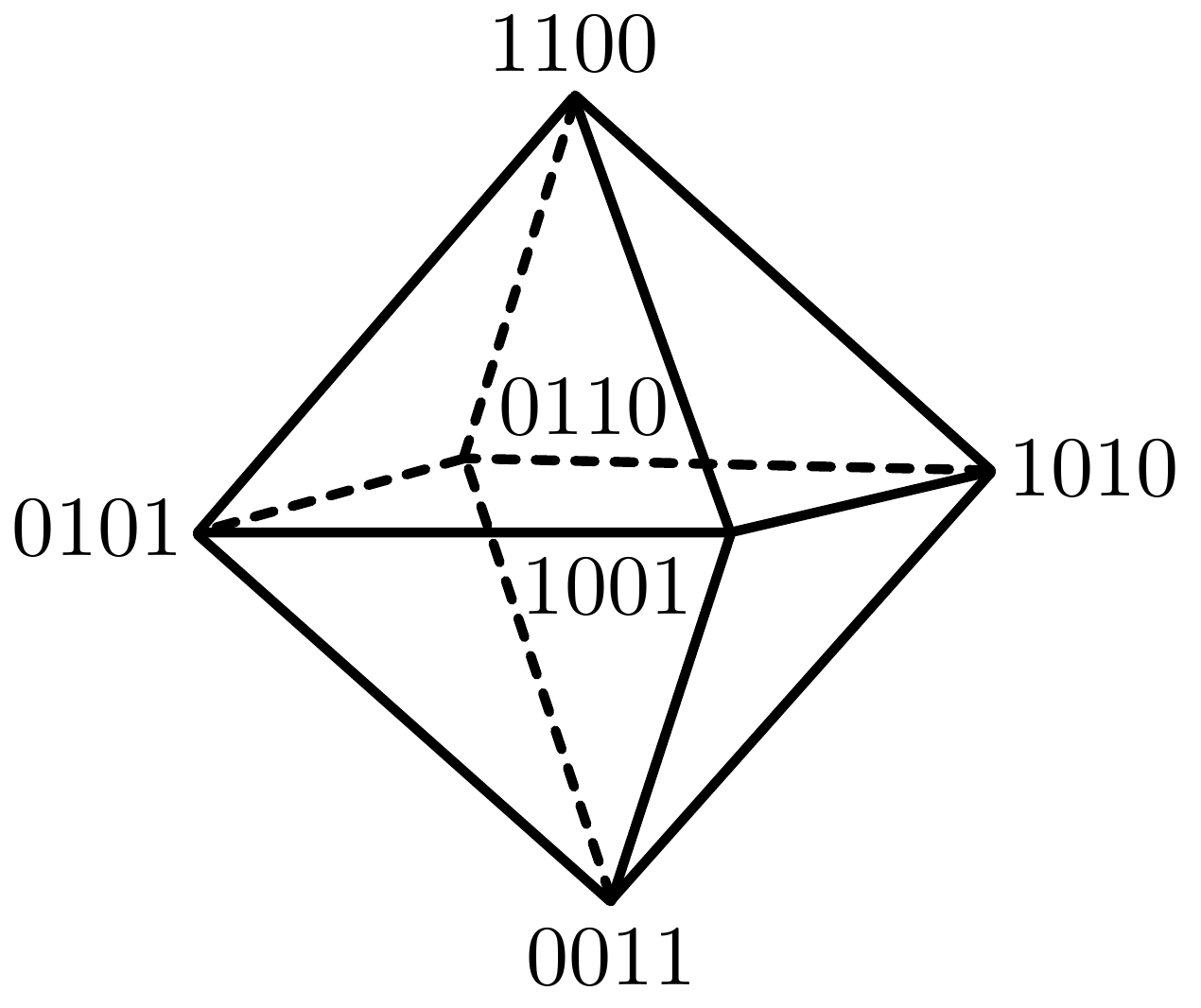}
  \caption{
    The \emph{limited multi-label} (LML) polytope $\LML_{n,k}$ from
    \citet{amos2019limited}
    is the set of points in the unit
    $n$-hypercube with coordinates that sum to $k$.
    $\LML_{n,1}$ is the \mbox{$(n-1)$-simplex}.
    The $\LML_{3,1}$ and $\LML_{3,2}$ polytopes (triangles) are on the left
    in blue. The $\LML_{4,2}$ polytope (an octahedron) is on the right.
    This polytope is also referred to as the \emph{knapsack polytope}
    or \emph{capped simplex}.
  }
  \label{fig:lml-example}
\end{figure}

The top-$k$ operation can be made differentiable by
replacing it with a soft version
\citep{martins2017learning,malaviya2018sparse,amos2019limited},
or by using a stochastic oracle \citep{brookes2018design}.
Here we use the
\emph{Limited Multi-Label Projection (LML) layer} \citep{amos2019limited},
which is a \emph{Bregman projection} of
points from $\sR^n$ onto
the \emph{LML polytope}
shown in \cref{fig:lml-example} and defined by
\begin{equation}
  \LML_{n,k} \defeq \{p\in\sR^n \mid 0\leq p\leq 1 \;\;
    {\rm and} \;\; 1^\top p = k\}.
\end{equation}
The LML polytope is the set of points in the unit
$n$-hypercube with coordinates that sum to $k$ and
is useful for modeling in multi-label and top-$k$ settings.
If $n$ is implied by the context we will
leave it out and write $\LML_k$.

We propose a \emph{temperature-scaled} LML variant
to project onto the interior of the LML polytope with
\begin{equation}
  \Pi_{\LML_k}(\nicefrac{x}{\tau}) \defeq \argmin_{0<y<1} \;\; -x^\top y - \tau H_b(y) \;\; \st\;\; 1^\top y = k
  \label{eq:lml-proj}
\end{equation}
where $\tau>0$ is the temperature parameter and
$$H_b(y) \defeq - \sum_i y_i\log y_i + (1-y_i)\log (1-y_i)$$
is the binary entropy function.
We introduce the hyper-parameter $\tau$ to show how DCEM
captures CEM as a special case as $\tau\rightarrow 0$.
\Cref{eq:lml-proj} is a convex optimization layer and
can be solved in a negligible amount of time with a GPU-amenable
bracketing method on the univariate dual as described
in \citet{amos2019limited}.
The derivative $\nabla_x \Pi_{\LML_k}(\nicefrac{x}{\tau})$
necessary for backpropagation can be easily computed by
implicitly differentiating the KKT optimality conditions as
described in \citet{amos2019limited}.

We can use the LML layer to make a soft and differentiable
version of \cref{eq:cem-update} as
\begin{equation}
  \begin{split}
    \phi_{t+1} \defeq \argmax_\phi &\;\; \sum_i \gI_{t,i} \log g_\phi(X_{t,i}) \\
    \subjectto &\;\; \gI_t = \Pi_{\LML_k}(\nicefrac{v_t}{\tau}).
  \end{split}
  \label{eq:cem-update-soft}
\end{equation}
This is now a
\emph{maximum weighted likelihood} estimation problem
\citep{markatou1997weighted,markatou1998weighted,wang2001maximum,hu2002weighted},
which still admits an analytic closed-form solution in
many cases, \eg for the natural exponential family \citep{de2005tutorial}.
Thus using the soft top-$k$ operation with the reparameterization trick, \eg,
on the samples from $g$ results in a differentiable variant of
CEM that we call DCEM and summarize in \cref{alg:dcem}.
We usually also normalize the values in each iteration
to help separate the scaling of the values from the temperature parameter.

\begin{algorithm*}[t]
  \caption{DCEM($f_\theta, g_\phi, \phi_1; \tau, N, k, T$)}
\label{alg:dcem}
\begin{algorithmic}
  \State DCEM minimizes a parameterized objective function $f_\theta$
  and is differentiable w.r.t.~$\theta$.
  Each DCEM iteration samples from the distribution $g_\phi$,
  starting with $\phi_1$.
  DCEM enables the derivative of $\E[g_{\phi_{T+1}}(\cdot)]$
  with respect to $\theta$ to be computed by differentiating
  all of the iterative operations.
  \vspace{1mm}\hrule\vspace{1mm}
  \For{t = 1 to T}
  \State $\left[X_{t,i}\right]_{i=1}^N \sim g_{\phi_t}(\cdot)$
  \Comment Sample $N$ points from the domain. \emph{Differentiate with reparameterization.}
  \State $v_{t,i}=f_\theta(X_{t,i})$
  \Comment Evaluate the objective function at those points.
  \State $\gI_t = \Pi_{\LML_k}(\nicefrac{v_t}{\tau})$
  \Comment Compute the soft top-$k$ projection of the values with \cref{eq:lml-proj}. \emph{Implicitly differentiate}
  \State Update $\phi_{t+1}$ by solving the maximum weighted likelihood problem in \cref{eq:cem-update-soft}.
  \EndFor
  \State \Return $\E[g_{\phi_{T+1}}(\cdot)]$
\end{algorithmic}
\end{algorithm*}

\begin{proposition}
  The temperature-scaled LML layer $\Pi_{\LML_k}(\nicefrac{x}{\tau})$
  approaches the hard top-$k$ operation as $\tau\rightarrow 0^+$
  when all components of $x$ are unique.
  \label{prop:temp-lml}
\end{proposition}

We prove this in \cref{app:temp-lml-proof} with
the KKT conditions of \cref{eq:lml-proj}.
The only difference between CEM and DCEM is the soft top-$k$
operation, thus when the soft top-$k$ operation approaches
the hard top-$k$ operation, we can conclude:

\begin{corollary}
  DCEM becomes CEM as $\tau\rightarrow 0^+$.
\end{corollary}

\begin{proposition}
  With an isotropic Gaussian sampling distribution, the
  maximum weighted likelihood update in
  \cref{eq:cem-update-soft} becomes
  $\mu_{t+1}\defeq \nicefrac{1}{k} \sum_i \gI_{t,i} X_{t,i}$
  and $\sigma^2_{t+1}\defeq \nicefrac{1}{k} \sum_{i} \gI_{t,i} \left(X_{t,i}-\mu_{t+1}\right)^2$,
  where the soft top-$k$ indexing set is
  $\gI_t \defeq \Pi_{\LML_k}(\nicefrac{v_t}{\tau})$.
  \label{prop:cem-wle}
\end{proposition}

This is well-known and is discussed in, \eg, \citet{de2005tutorial},
and can be proved by differentiating \cref{eq:cem-update-soft}.

\begin{corollary}
\Cref{prop:cem-wle} captures
\cref{prop:vanilla-cem} as $\tau\rightarrow 0^+$.
\end{corollary}

\section{Applications}
\subsection{Energy-Based Learning}
\label{sec:app:eb}

\emph{Energy-based learning} for regression and classification
estimate the conditional probability $\sP(y|x)$ of an output $y\in\gY$
given an input $x\in\gX$ with a parameterized energy function
\mbox{$E_\theta(y|x)\in\gY\times\gX \rightarrow \sR$}
such that $\sP(y|x) \propto \exp\{-E_\theta(y|x)\}$.
\emph{Predictions} are made by solving the optimization problem
\begin{equation}
  \hat y \defeq \argmin_y E_\theta(y|x).
  \label{eq:spen-inference}
\end{equation}
Historically linear energy functions have been well-studied,
\eg in \citet{taskar2005learning,lecun2006tutorial},
as it makes \cref{eq:spen-inference} easier to solve
and analyze.
More recently non-convex energy functions that are parameterized
by neural networks are being explored --- a popular one being
\emph{Structured Prediction Energy Networks} (SPENs)
\citep{belanger2016structured} which propose to model
$E_\theta$ with neural networks.
\citet{belanger2017end} does supervised learning of
SPENs by approximating \cref{eq:spen-inference}
with gradient descent that is then unrolled for $T$ steps,
\ie by starting with some $y_0$, making
gradient updates
$$y_{t+1}\defeq y_t-\gamma \nabla_y E_\theta(y_t|x)$$
resulting in an output $\hat y \defeq y_T$,
defining a loss function $\Ls$ on top of $\hat y$,
and doing learning with gradient updates $\nabla_\theta \Ls$
that go through the inner gradient steps.

In this context we can alternatively use DCEM to approximate
\cref{eq:spen-inference}.
One potential consideration when training deep energy-based
models with approximations to \cref{eq:spen-inference} is
the impact and bias that the approximation is going to have
on the energy surface.
We note that for gradient descent, \eg, it may cause the energy
surface to overfit to the number of gradient steps so that
the output of the approximate inference procedure isn't even
a local minimum of the energy surface.
One potential advantage of DCEM is that the output is more likely
to be near a local minimum of the energy surface so that,
\eg, more test-time iterations can be used to refine the solution.
We empirically illustrate the impact of the optimizer choice
on a synthetic example in \cref{sec:exp:reg}.

\begin{algorithm*}[t]
  \caption{Learning an embedded control space with DCEM}
\label{alg:embed}
\begin{algorithmic}
  \State \textbf{Fixed Inputs:} Dynamics $\ftrans$,
  per-step state-action cost $C_t(x_t, u_t)$ that induces $C_\theta(z; \xinit)$,
  horizon $H$,
  full control space $\gU^H$, distribution over initial states $\gD$
  \State \textbf{Learned Inputs:} Decoder $\fdec_\theta: \gZ \rightarrow \gU^H$
  \vspace{1mm}\hrule\vspace{1mm}
  \While{not converged}
  \State Sample initial state $\xinit\sim \gD$.
  \State $\hat z = \argmin_{z\in \gZ} C_\theta(z; \xinit)$
  \Comment Solve the embedded control problem \cref{eq:emb_ctrl}
  with DCEM.
  \State $\theta \leftarrow \text{grad-update}(\nabla_\theta C_\theta(\hat z))$
  \Comment Update the decoder to improve the controller's cost.
  \EndWhile
\end{algorithmic}
\end{algorithm*}

\subsection{Control and Reinforcement Learning}
\label{sec:app:ctrl}

Our main application focus is in the \emph{continuous control} setting
where we show how to use DCEM to learn a \emph{latent control space} that
is easier to solve than the original problem \emph{and} induces
a \emph{differentiable policy class} that allows parts of the controller
to be fine-tuned with auxiliary policy or imitation losses.

We are interested in controlling \emph{discrete-time} dynamical
systems with \emph{continuous} state-action spaces.
Let $H$ be the \emph{horizon length} of the controller
and $\gU^H$ be the space of control sequences over this
horizon length, \eg
$\gU$ could be a multi-dimensional real space or box therein
and $\gU^H$ could be the Cartesian product of those spaces
representing the sequence of controls over $H$ timesteps.
We are interested in repeatedly solving the control
optimization problem\footnote{%
  We omit some explicit
  variables from the $\argmin$ operator when they are
  can be inferred by the context.}
\begin{equation}
  \begin{split}
    \hat u_{1:H} \defeq \argmin_{u_{1:H}\in \gU^H} &\;\; \sum_{t=1}^H  C_t(x_t, u_t) \\
    \subjectto &\;\; x_1 = \xinit \\
    &\;\; x_{t+1} = \ftrans(x_t, u_t)
  \end{split}
  \label{eq:ctrl}
\end{equation}
where we are in an initial system state $\xinit$
governed by deterministic system \emph{transition dynamics} $\ftrans$,
and wish to find the optimal sequence of actions $\hat u_{1:H}$
such that we find a valid trajectory $\{x_{1:H}, u_{1:H}\}$
that optimizes the cost $C_t(x_t, u_t)$.
\Cref{eq:ctrl} can be seen as an instance of
\cref{eq:intro_argmin} by moving the rollout of the
dynamics into the cost function.
Typically these controllers are used for \emph{receding horizon} control
\citep{mayne1990receding}
where only the first action $u_1$
is deployed on the real system, a new state is obtained from the system,
and the \cref{eq:ctrl} is solved again from the new initial state.
In this case we can say the controller induces a \emph{policy}
$\pi(\xinit)\defeq\hat u_1$\footnote{%
  We also omit the dependency of $u_1$ on $\xinit$.}
that solves \cref{eq:ctrl}
and depends on the cost and transition dynamics, and potential
parameters therein.
In all of the cases we consider $\ftrans$ is deterministic, but
may be approximated by a stochastic model for learning.
Some \emph{model-based} reinforcement learning settings consider
cases where $\ftrans$ and $C$ are parameterized and potentially
used in conjunction with another policy class.

For sufficiently complex dynamical systems, \cref{eq:ctrl}
is computationally expensive and numerically instable to solve
and rife with sub-optimal local minima.
The cross-entropy method is the state-of-the-art method for
solving \cref{eq:ctrl} with neural network
transitions $\ftrans$ \citep{chua2018deep,hafner2018learning}.
CEM in this context samples full action sequences and refines
the samples towards ones that solve the control problem.
\citet{hafner2018learning} uses CEM with 1000 samples in each
iteration for 10 iterations with a horizon length of 12.
This requires $1000\times 10\times 12=120,000$ evaluations (!) of
the transition dynamics to predict the control to be taken
given a system state --- and the transition dynamics may use
a deep recurrent architecture as in \citet{hafner2018learning}
or an ensemble of models as in \citet{chua2018deep}.
One comparison point here is a model-free neural network policy
takes a \emph{single} evaluation for this prediction,
albeit sometimes with a larger neural network.

The first application we show of DCEM in the continuous control
setting is to learn a \emph{latent action space} $\gZ$
with a parameterized \emph{decoder} $\fdec_\theta: \gZ\rightarrow\gU^H$
that maps back up to the space of \emph{optimal} action \emph{sequences},
which we illustrate in \cref{fig:rssm}.
For simplicity starting out, assume that the dynamics and cost
functions are known (and perhaps even the ground-truth)
and that the only problem is to estimate the decoder in isolation,
although we will show later that these assumptions can be relaxed.
The motivation for having such a latent space and decoder is
that the millions of times \cref{eq:ctrl} is being solved
for the same dynamic system with the same cost,
the solution space of \emph{optimal} action sequences
$\hat u_{1:H}\in\gU^H$
has an extremely large amount of \emph{spatial} (over $\gU$)
and \emph{temporal} (over time in $\gU^H$) structure that
is being ignored by CEM on the full space.
The space of optimal action sequences only contains the
knowledge of the trajectories that matter for solving the
task at hand, such as different parts of an optimal gait,
and not irrelevant control sequences.
We argue that CEM over the full action space wastes a lot
of computation considering irrelevant action sequences and
show that these can be ignored by learning a latent space
of more reasonable candidate solutions here that
we search over instead.
Given a decoder, the control optimization problem in
\cref{eq:ctrl} can then be transformed into an optimization
problem over $\gZ$ as
\begin{equation}
  \begin{split}
    \hat z \defeq \argmin_{z\in \gZ} \;\; & C_\theta(z; \xinit) \defeq\sum_{t=1}^H  C_{t}(x_t, u_t) \\
    \;\; \subjectto \;\; & x_1 = \xinit \\
    & x_{t+1} = \ftrans(x_t, u_t) \\
    & u_{1:H}=\fdec_\theta(z)
  \end{split}
  \label{eq:emb_ctrl}
\end{equation}
which is still a challenging non-convex optimization problem
that searches over a decoder's input space to find the
optimal control sequence.
\Cref{eq:emb_ctrl} can be seen as an instance of
\cref{eq:intro_argmin} by moving the decoder and
rollout of the dynamics into the cost function $C_\theta(z; \xinit)$
and can be solved with CEM and DCEM.
We notationally leave out the dependence of $\hat z$
on $\xinit$ and $\theta$.

\begin{figure*}[t]
  \centering
  \includegraphics[width=0.8\textwidth]{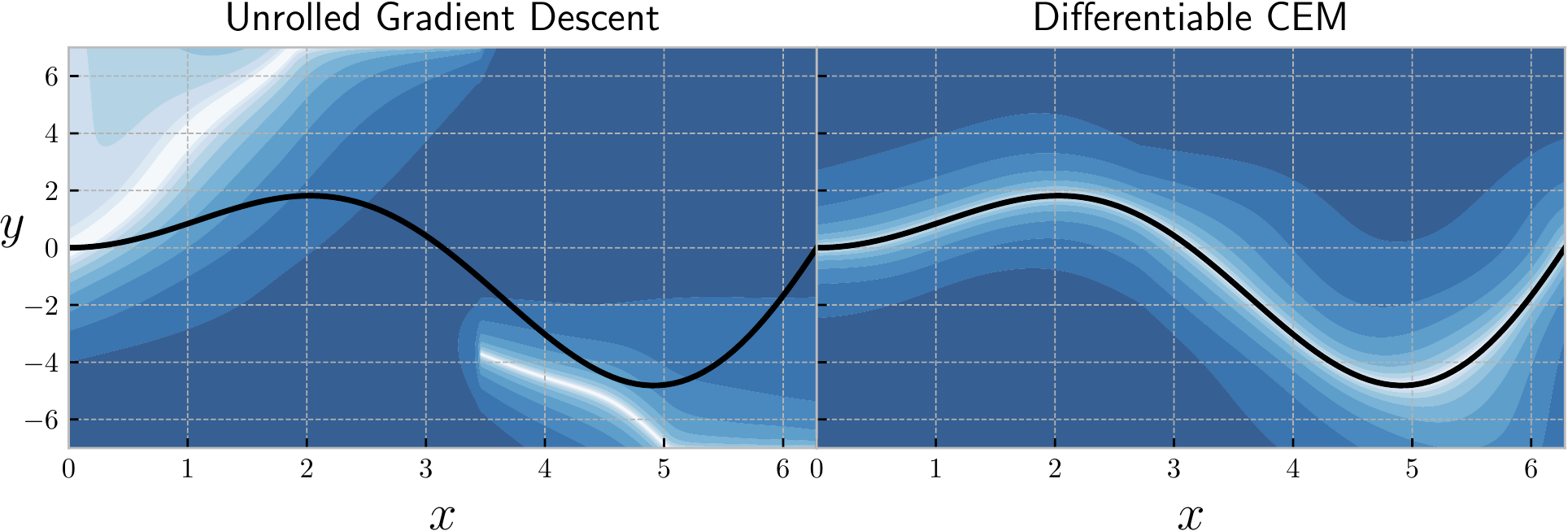}
  \caption{
    We trained an energy-based model with unrolled gradient descent
    and DCEM for 1D regression onto the black target function.
    Each method unrolls through 10 optimizer steps.
    The contour surfaces show the (normalized/log-scaled)
    energy surfaces, highlighting that unrolled gradient descent models
    can overfit to the number of gradient steps.
    The lighter colors show areas of lower energy.
  }
  \label{fig:exp:reg:e-surfaces}
\end{figure*}

We propose in \cref{alg:embed}
to use DCEM to approximately solve \cref{eq:emb_ctrl}
and then \emph{learn} the decoder directly to optimize the
performance of \cref{eq:ctrl}.
Every time we solve \cref{eq:emb_ctrl} with DCEM and obtain
an optimal latent representation $\hat z$ along with the
induced trajectory $\{x_t, u_t\}$, we can take a
gradient step to push down the resulting cost of
that trajectory with $\nabla_\theta C(\hat z)$, which
goes through the DCEM process that uses the decoder
to generate samples to obtain $\hat z$.
The DCEM machinery behind this is not necessary \emph{if}
a reasonable local minima is consistently found as this
is an instance of min-differentiation
\citep[Theorem~10.13]{rockafellar2009variational}
but in practice this breaks down in non-convex cases
when the minimum cannot be consistently found.
DCEM helps by providing the derivative information
$\nabla_\theta \hat z$.
\citet{antonova2019bayesian,wang2019exploring}
solve related problems in this space and we discuss
them in \cref{ref:bg:rl}.
Learning an action embedding also requires derivatives through
the transition dynamics and cost functions
to compute $\nabla_\theta C(\hat z)$,
even if the ground-truth dynamics are being used.
This gives the latent space the knowledge of how
the control cost will change as the decoder's parameters change.

DCEM in this setting also induces a \emph{differentiable policy class}
$\pi(\xinit)\defeq u_1=\fdec(\hat z)_1$.
This enables a policy or imitation loss $\gJ$
to be defined on the policy that can fine-tune the parts
of the controller (decoder, cost, and transition dynamics)
gradient information from $\nabla_\theta \gJ$.
In theory the same approach could be used with CEM on the full
optimization problem in \cref{eq:ctrl}.
For realistic problems without modification this is intractable
and memory-intensive as it would require storing and
backpropagating through every sampled trajectory, although
as a future direction we note that it may be possible to delete
some of the low-influence trajectories to help overcome this.

\section{Experiments}
Our experiments demonstrate applications of the cross-entropy method
in structured prediction, control, and reinforcement learning.
\cref{sec:exp:reg} illustrate a synthetic regression structured
prediction task where gradient descent learns a counter-intuitive
energy surface while DCEM retains the minimum.
\cref{sec:ctrl} shows how DCEM can embed control optimization problems
in a case when the ground-truth model is known or unknown, and
we show that PPO \citep{schulman2017proximal} can help improve
the embedded controller.

Our PyTorch \citep{paszke2019pytorch} source code is openly available at
\href{http://github.com/facebookresearch/dcem}{github.com/facebookresearch/dcem}
and uses the PyTorch LML implementation
from \href{https://github.com/locuslab/lml}{github.com/locuslab/lml}
to compute \cref{eq:lml-proj}.

\subsection{Unrolling optimizers for regression
  and structured prediction}
\label{sec:exp:reg}

In this section we briefly explore the impact of the inner
optimizer on the energy surface of a SPEN as discussed in
\cref{sec:app:eb}.
For illustrative purposes we consider a simple unidimensional regression task
where the \emph{ground-truth data} is generated from $f(x)\defeq x\sin(x)$
for $x\in[0, 2\pi]$.
We model $\sP(y|x) \propto \exp\{-E_\theta(y|x)\}$
with a single neural network $E_\theta$ and make predictions
$\hat y$ by solving the optimization problem
\cref{eq:spen-inference}.
Given the ground-truth output $y^\star$,
we use the loss $\Ls(\hat y, y^\star) \defeq ||\hat y - y^\star||_2^2$
and take gradient steps of this loss to shape the energy landscape.

\begin{figure*}[t]
  \centering
  \includegraphics[width=\textwidth]{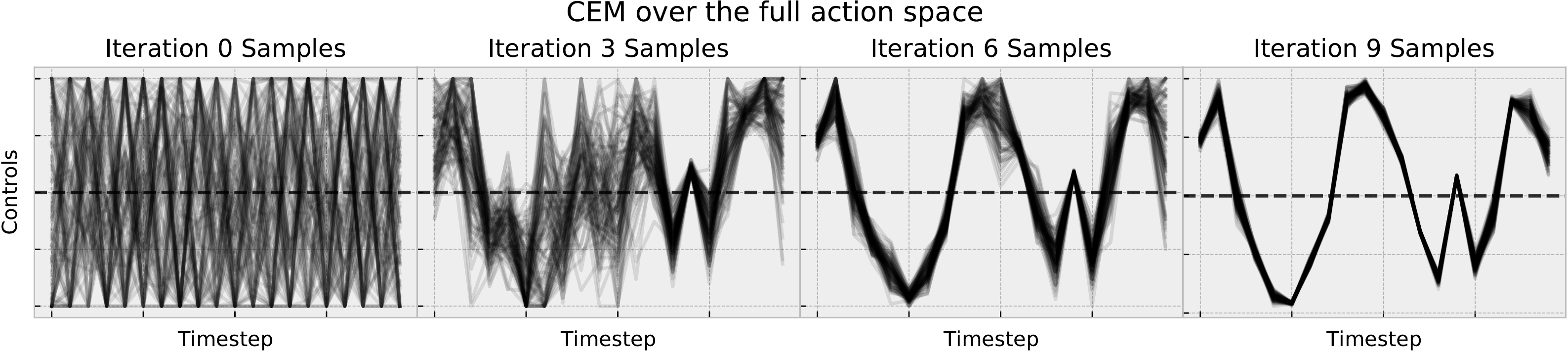} \\[4mm]
  \includegraphics[width=\textwidth]{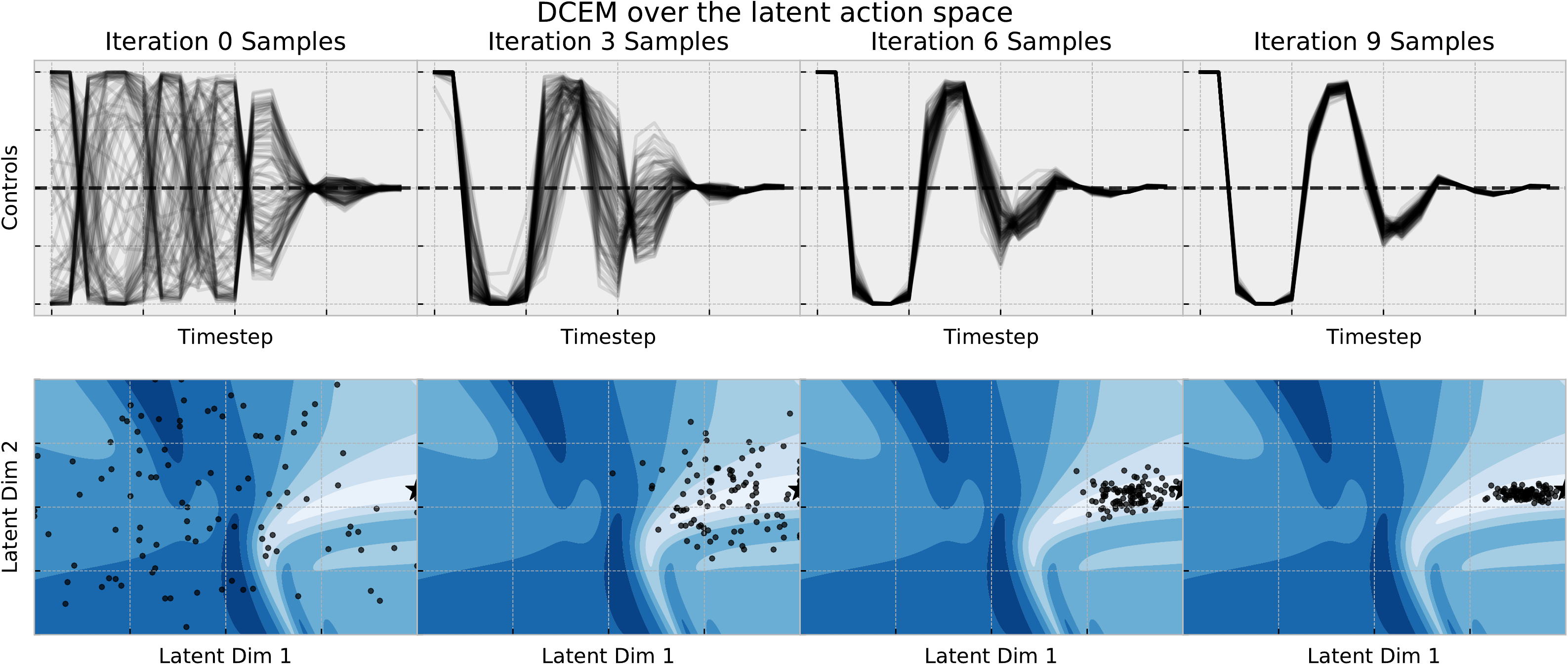}
  \caption{
    Visualization of the samples that CEM and DCEM generate
    to solve the cartpole task starting from the same initial
    system state.
    The plots starting at the top-left show that CEM initially starts with
    no temporal knowledge over the control space whereas embedded
    DCEM's latent space generates a more feasible distribution
    over control sequences to consider in each iteration.
    Embedded DCEM uses an order of magnitude less samples and is
    able to generate a better solution to the control problem.
    The contours on the bottom show the controller's cost
    surface $C(z)$ from \cref{eq:emb_ctrl} for
    the initial state --- the lighter colors show
    regions with lower costs.
  }
  \label{fig:emb:cp}
\end{figure*}

We consider approximating \cref{eq:spen-inference} with unrolled
gradient descent and DCEM with Gaussian sampling distributions.
Both of these are trained to take 10
optimizer steps and we use an inner learning rate of 0.1 for
gradient descent and with DCEM we use 10 iterations with
100 samples per iteration and 10 elite candidates, with
a temperature of 1.
For both algorithms we start the initial iterate at $y_0\defeq 0$.
We show in \cref{app:reg} that both of these models attain the
same loss on the training dataset but, since this is a unidimensional
regression task, we can visualize the entire energy surfaces
over the joint input-output space in \cref{fig:exp:reg:e-surfaces}.
This shows that gradient descent has learned to adapt from the
initial $y_0$ position to the final position by descending
along the function's surface as we would expect, but there
is no reason why the energy surface should be a local minimum
around the last iterate $\hat y \defeq y_{10}$.
The energy surface learned by CEM captures local minima around
the regression target as the sequence of Gaussian iterates
are able to capture a more global view of the function landscape
and need to focus in on a minimum of it for regression.
We show ablations in \cref{app:reg} from training for 10 inner
iterations and then evaluating with a different number of iterations
and show that gradient descent quickly steps away from
making reasonable predictions.

\textbf{Discussion.}
Other tricks \emph{could} be used to force the output
to be at a local minimum with gradient descent, such as
using multiple starting points or randomizing the number of
gradient descent steps taken --- our intention here is to
highlight this behavior in the vanilla case.
DCEM is also susceptible to overfitting to the
hyper-parameters behind it in similar, albeit less obvious ways.

\subsection{Control}
\label{sec:ctrl}
\subsubsection{Starting simple: Embedding the cartpole's action space}
\label{sec:ctrl:cp}
We first show that it is possible to learn an embedded control
space as discussed in \cref{sec:app:ctrl} in an isolated setting.
We use the standard cartpole dynamical system from
\citet{barto1983neuronlike} with a \emph{continuous} state-action space.
We assume that the ground-truth dynamics and cost are \emph{known}
and use the differentiable ground-truth dynamics and cost implemented
in PyTorch from \citet{amos2018differentiable}.
This isolates the learning problem to \emph{only} learning
the embedding so that we can study what this is doing
without the additional complications that arise from
exploration, estimating the dynamics, learning a policy,
and other non-stationarities.
We show experiments with these assumptions relaxed in \cref{sec:ctrl:dmc}.

We use DCEM and \cref{alg:embed}
to learn a 2-dimensional latent space $\gZ\defeq [0,1]^2$ that maps
back up to the full control space $\gU^H\defeq [0,1]^H$ where we
focus on horizons of length $H\defeq 20$.
For DCEM over the embedded space we use 10 iterations with
100 samples in each iteration and 10 elite candidates,
again with a temperature of 1.
We show the details in \cref{app:cartpole} that we \emph{are} able to
recover the performance of an expert CEM controller that
uses an order-of-magnitude more samples
\cref{fig:emb:cp} shows a visualization of what the CEM
and embedded DCEM iterates look like to solve the control
optimization problem from the same initial system state.
CEM spends a lot of evaluations on sequences in the control
space that are unlikely to be optimal, such as the ones the
bifurcate between the boundaries of the control space at
every timestep, while our embedded space
is able to learn more reasonable proposals.

\subsubsection{Scaling up to continuous locomotion}
\label{sec:ctrl:dmc}

\begin{figure*}[t]
  \centering
  \begin{minipage}{0.24\linewidth}
    \raggedleft
    \vspace{-2mm}
    Full CEM \\[5mm]
    Latent DCEM \\[5mm]
    Latent DCEM+PPO
  \end{minipage}%
  \vspace{2mm}
  \begin{minipage}{0.37\linewidth}
  \includegraphics[width=0.8\textwidth]{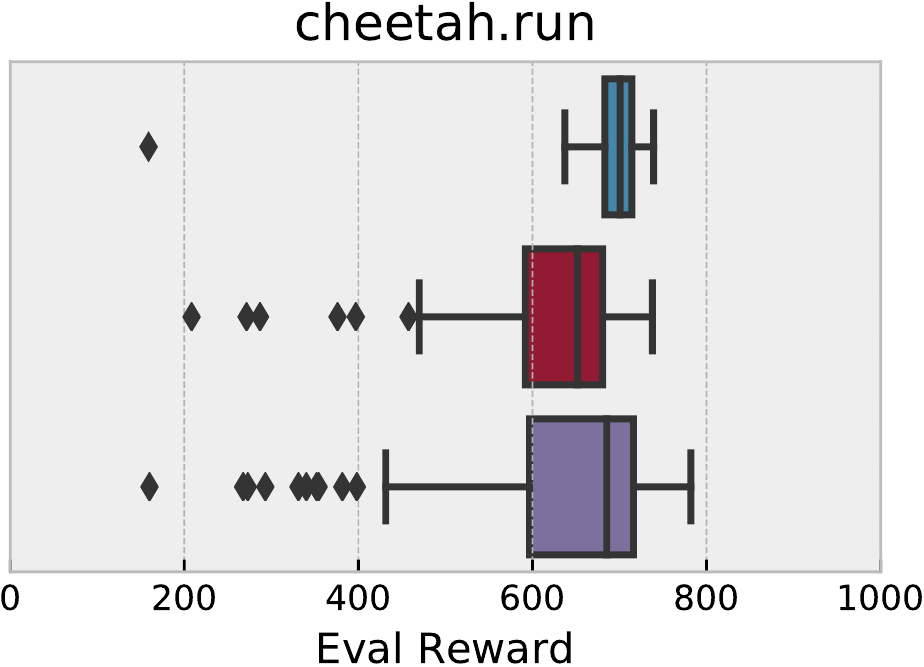}
  \end{minipage}%
  \begin{minipage}{0.37\linewidth}
  \includegraphics[width=0.8\textwidth]{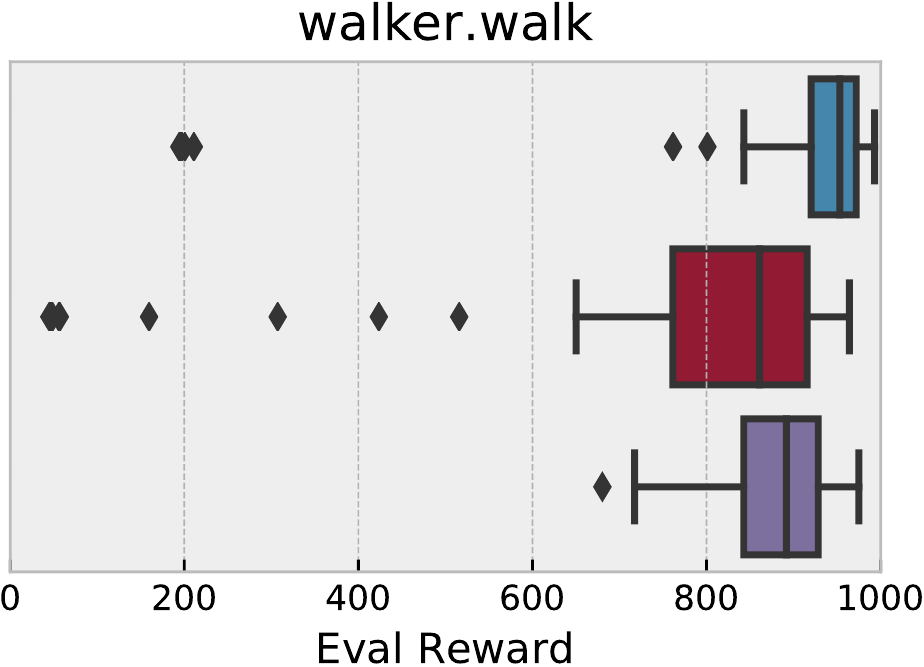}
  \end{minipage}%
  \caption{
    We evaluated our final models by running 100 episodes each
    on the cheetah and walker tasks.
    CEM over the full action space uses 10,000 trajectories
    for control at each time step while embedded DCEM
    samples only 1000 trajectories.
    DCEM almost recovers the performance of CEM over the
    full action space and PPO fine-tuning of the model-based
    components helps bridge the gap.
  }
  \label{fig:dmc:rew}
\end{figure*}

Next we show that we can relax the assumptions of having
known transition dynamics and reward and show that we can
learn a latent control space on top of a learned model
on the \verb!cheetah.run! and \verb!walker.walk!
continuous locomotion tasks from the DeepMind control
suite \citep{tassa2018deepmind} using the MuJoCo
physics engine \citep{todorov2012mujoco}.
We then fine-tune the policy induced by the embedded
controller with PPO \citep{schulman2017proximal},
sending the policy loss directly back into the reward
and latent embedding modules underlying the controller.
Videos of our trained models are available at
\href{https://sites.google.com/view/diff-cross-entropy-method}{https://sites.google.com/view/diff-cross-entropy-method}.

We start with a state-of-the-art model-based RL approach
by noting that the PlaNet \citep{hafner2018learning}
restricted state space model (RSSM) is a reasonable
architecture for proprioceptive-based control in addition
to just pixel-based control.
We show the graphical model we use in \cref{fig:rssm},
which maintains deterministic hidden states
$h_t$ and stochastic (proprioceptive) system observations
$x_t$ and rewards $r_t$.
We model transitions as $h_{t+1}=\ftrans_\theta(h_t, x_t)$,
observations with $x_t\sim \fodec_\theta(h_t)$,
rewards with $r_t = \frew_\theta(h_t, x_t)$,
and map from the latent action space to action sequences
with $u_{1:T}=\fdec(z)$.
We follow the online training procedure of \citet{hafner2018learning}
to initialize all of the models except for the action
decoder $\fdec$, using approximately 2M timesteps.
We then use a variant of \cref{alg:embed} to learn $\fdec$ to
embed the action space for control with DCEM, which we also
do online while updating the models.
We describe the full training process in
\cref{app:control-suite}.

Our DCEM controller induces a differentiable policy class
$\pi_\theta(\xinit)$ where $\theta$ are the parameters of
the models that impact the actions that the controller
is selecting.
We then use PPO to define a loss on top of this policy
class and fine-tune the components (the decoder and reward module)
so that they improve the episode reward rather than
the maximum-likelihood solution of observed trajectories.
We chose PPO because we thought it would be able to fine-tune
the policy with just a few updates because the policy
is starting at a reasonable point, but this did not
turn out to be the case and in the future other policy
optimizers can be explored.
We implement this by making our DCEM controller the
policy in the PPO implementation by \citet{pytorchrl}.
We provide more details behind our training procedure in
\cref{app:control-suite}.

We evaluate our controllers on 100 test episodes and
the rewards in \cref{fig:dmc:rew} show that DCEM
is almost (but not exactly) able to recover the
performance of doing CEM over the full action space
while using an order-of-magnitude less trajectory samples
(1,000 vs 10,0000).
PPO fine-tuning helps bridge the gap between the
performances.

\textbf{Discussion.}
The future directions of DCEM in the control setting
will help bring efficiency and policy-based
fine-tuning to model-based reinforcement learning.
Much more analysis and experimentation is necessary to
achieve this as we faced many issues getting the
model-based cheetah and walker tasks to work that
did not arise in the ground-truth cartpole task.
We discuss this more in \cref{app:control-suite}.
We also did not focus on the sample complexity of our
algorithms getting these proof-of-concept experiments working.
Other reasonable baselines on
this task could involve distilling the
controller into a model-free policy and
then doing search on top of that policy,
as done in POPLIN \citep{wang2019exploring}.

\section{Conclusions and Future Directions}
We have shown how to differentiate through the
cross-entropy method and have brought CEM into the
end-to-end learning pipeline.
Beyond further explorations in the energy-based learning and control contexts
we showed here, DCEM can be used anywhere gradient descent is unrolled.
We find this especially promising for meta-learning and
can build on LEO \citep{rusu2018meta} or CAVIA \citep{zintgraf2019fast}.
Inspired by DCEM, other more powerful sampling-based
optimizers could be made differentiable in the same way,
potentially optimizers that leverage gradient-based information
in the inner optimization steps
\citep{sekhon1998genetic,theodorou2010generalized,stulp2012path,maheswaranathan2018guided}
or by also learning the hyper-parameters of structured optimizers
\citep{li2016learning,volpp2019meta,chen2017learning}.

\subsection*{Acknowledgments}
We thank
David Belanger,
Roberto Calandra,
Yinlam Chow,
Rob Fergus,
Mohammad Ghavamzadeh,
Edward Grefenstette,
Shubhanshu Shekhar,
and
Zolt{\'a}n Szab{\'o}
for insightful discussions,
and the anonymous reviewers for many useful suggestions
and improvements to this paper.

We acknowledge the Python community
\citep{van1995python,oliphant2007python}
for developing
the core set of tools that enabled this work, including
PyTorch \citep{paszke2019pytorch},
Hydra \citep{hydra},
Jupyter \citep{kluyver2016jupyter},
Matplotlib \citep{hunter2007matplotlib},
seaborn \citep{seaborn},
numpy \citep{oliphant2006guide,van2011numpy},
pandas \citep{mckinney2012python}, and
SciPy \citep{jones2014scipy}.

{\footnotesize
\bibliography{dcem}
\bibliographystyle{icml2020}
}

\newpage
\appendix
\section{Proof of \cref{prop:temp-lml}}
\label{app:temp-lml-proof}

\begin{proof}
  We first note that a solution exists to the projection operation,
  and it is unique, which comes from the strict convexity
  of the objective \citep{rao1984convexity}.
  The Lagrangian of the temperature-scaled LML projection in \cref{eq:lml-proj} is
  \begin{equation}
  L(y, \nu) = -x^\top y - \tau H_b(y) + \nu(k - 1^\top y).
  \label{eq:lag}
  \end{equation}
  Differentiating \cref{eq:lag} gives
  \begin{equation}
  \nabla_y L(y, \nu) = -x + \tau \log \frac{y}{1-y}-\nu
  \end{equation}
  and the first-order optimality condition
  $\nabla_y L(y^\star, \nu^\star)=0$ gives
  $y^\star_i = \sigma(\tau^{-1}(x_i+\nu^*))$,
  where $\sigma$ is the sigmoid function.
  Using \cref{lem:temp-sigmoid} as $\tau\rightarrow 0^+$ gives
  \begin{equation}
    y^\star_i =
    \begin{cases}
      1 & \text{if } x_i > -\nu^* \\
      0 & \text{if } x_i < -\nu^* \\
      \nicefrac{1}{2} & \text{otherwise.}
    \end{cases}
  \end{equation}
  Substituting this back into the constraint $1^\top y^\star = k$
  gives that $\pi(x)_k < -\nu^* < \pi(x)_{k+1}$,
  where
  $\pi(x)$ sorts $x\in\sR^n$ in ascending order so that
  $\pi(x)_1 \leq \pi(x)_2 \leq \ldots \leq \pi(x)_n.$
  Thus we have that $y^\star_i = \mathbbold{1}\{x_i\geq \pi(x)_k\}$,
  which is 1 when $x_i$ is in the top-$k$ components of $x$
  and 0 otherwise, and therefore the temperature-scaled LML layer
  approaches the hard top-$k$ function as
  $\tau\rightarrow 0^+$.
\end{proof}

\vspace{1cm}

\begin{lemma}
  \begin{equation}
    \lim_{\tau\rightarrow 0^+} \sigma(\nicefrac{x}{\tau}) =
    \begin{cases}
      1 & \text{if } x > 0 \\
      0 & \text{if } x < 0 \\
      \nicefrac{1}{2} & \text{otherwise,}
    \end{cases}
  \end{equation}
  where $\sigma(\nicefrac{x}{\tau})=(1+\exp\{\nicefrac{-x}{\tau}\})^{-1}$ is the
  temperature-scaled sigmoid.
  \label{lem:temp-sigmoid}
\end{lemma}

\section{More details: Simple regression task}
\label{app:reg}

\begin{figure*}[t]
  \centering
  \includegraphics[width=0.45\textwidth]{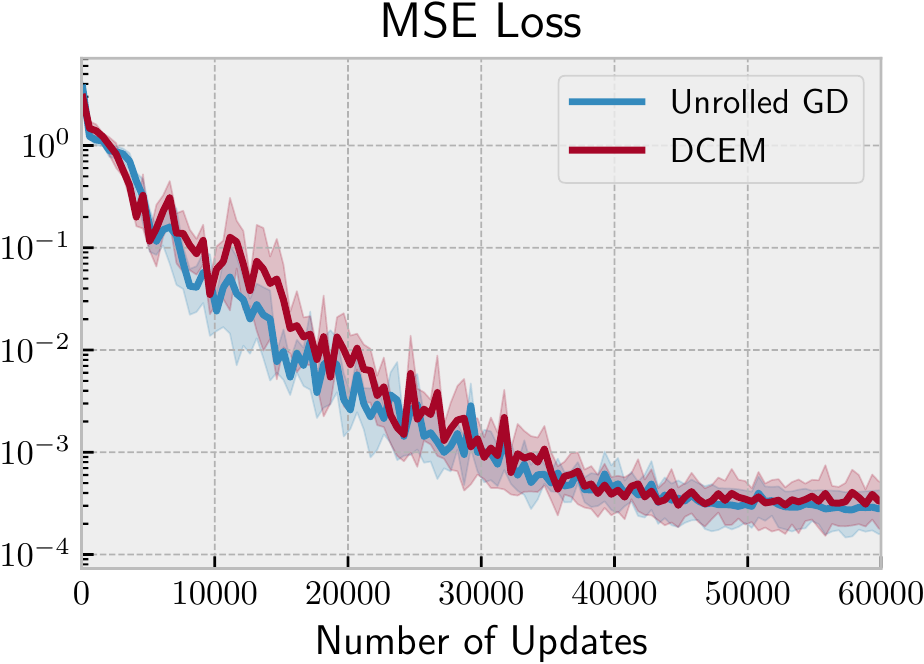}
  \includegraphics[width=0.45\textwidth]{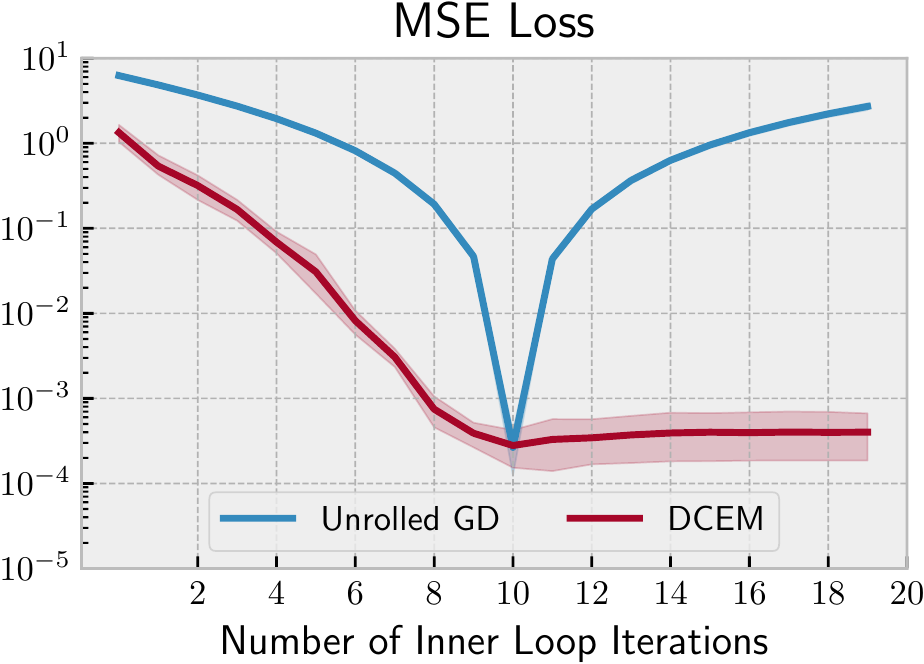}
  \caption{\textbf{Left:} Convergence of DCEM and unrolled GD on the
    regression task.
    \textbf{Right:} Ablation showing what happens when DCEM and unrolled GD are
    trained for 10 inner steps and then a different number of
    steps is used at test-time.
    We trained three seeds for each model and the shaded regions
    show the 95\% confidence interval.
  }
  \label{fig:reg:conv}
\end{figure*}

\begin{figure*}[t]
  \centering
  \includegraphics[width=0.8\textwidth]{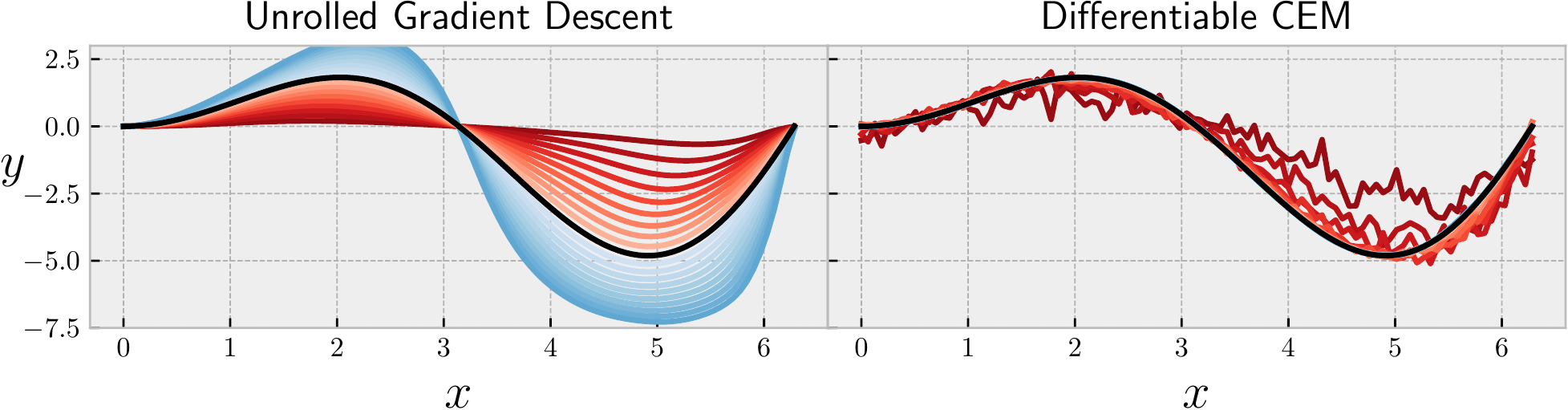} \\
  \begin{tikzpicture}
    \node[]{\includegraphics[width=0.7\textwidth]{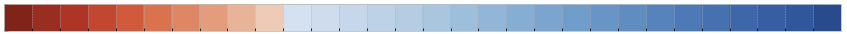}};
    \node[] at (0,0.27) {\tiny Number of Inner Loop Iterations};
    \node[anchor=west] (n1) at (-4.9,-0.27) {\tiny 1};
    \node[right=2.5 of n1,anchor=west] (n10) {\tiny 10};
    \node[right=2.71 of n10,anchor=west] (n20) {\tiny 20};
    \node[right=2.71 of n20,anchor=west] (n30) {\tiny 30};
  \end{tikzpicture}%
  \caption{
    Visualization of the predictions made by ablating the
    number of inner loop iterations for unrolled GD and DCEM.
    The ground-truth regression target is shown in black.
  }
  \label{fig:reg:pred-vis}
\end{figure*}
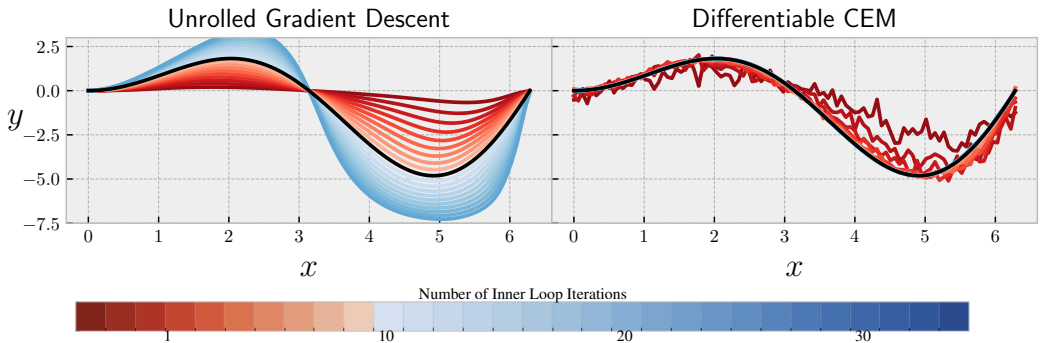

\Cref{fig:reg:conv} (left) shows the convergence of unrolled GD
and DCEM on the training data, showing that they are able
to obtain the same training loss despite inducing very
different energy surfaces.
\Cref{fig:reg:conv} (right) and \cref{fig:reg:pred-vis}
shows the impact of training gradient descent and
DCEM to take 10 inner optimization steps and then
ablating the number of inner steps at test-time.

\section{More details: Cartpole experiment}
\label{app:cartpole}

In this section we discuss some of the ablations we considered
when learning the latent action space for the cartpole task.
In all settings we use DCEM to unroll 10 inner iterations
that samples 100 candidate points in each iteration and has
an elite set of 10 candidates.

\begin{figure}[t]
  \centering
  \includegraphics[width=0.4\textwidth]{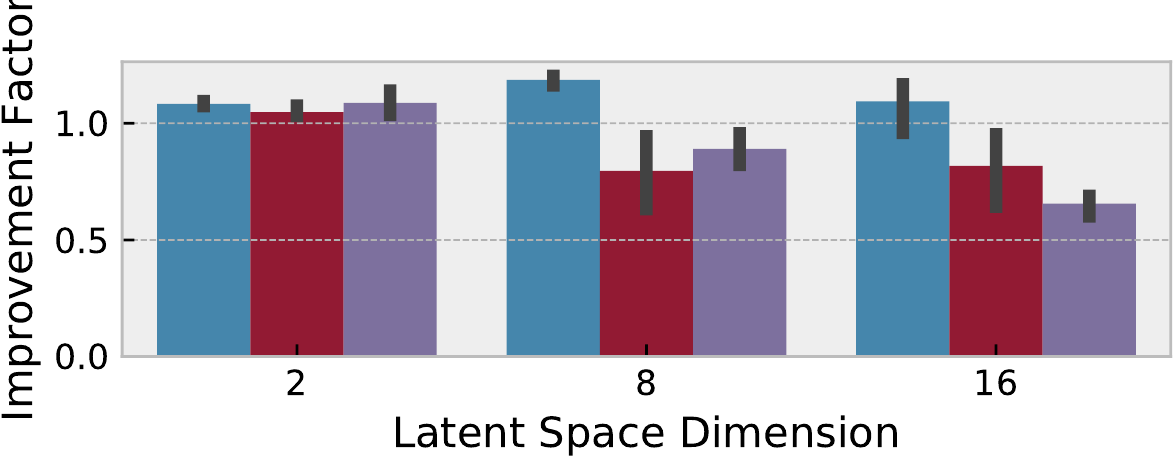} \\
  $\tau = $ (\cblock{83}{123}{164} 1.0 \cblock{187}{64}{60} 0.1 \cblock{159}{92}{149} 0.0)
  \caption{Improvement factor on the ground-truth cartpole task
    from embedding the action space with DCEM compared to
    running CEM on the full action space, showing that
    DCEM is able to recover the full performance.
    We use the DCEM model that achieves the best validation loss.
    The error lines show the 95\% confidence interval around three trials.
  }
  \label{fig:cp:improve}
\end{figure}

\begin{figure*}[ht]
  \centering
  \includegraphics[width=\textwidth]{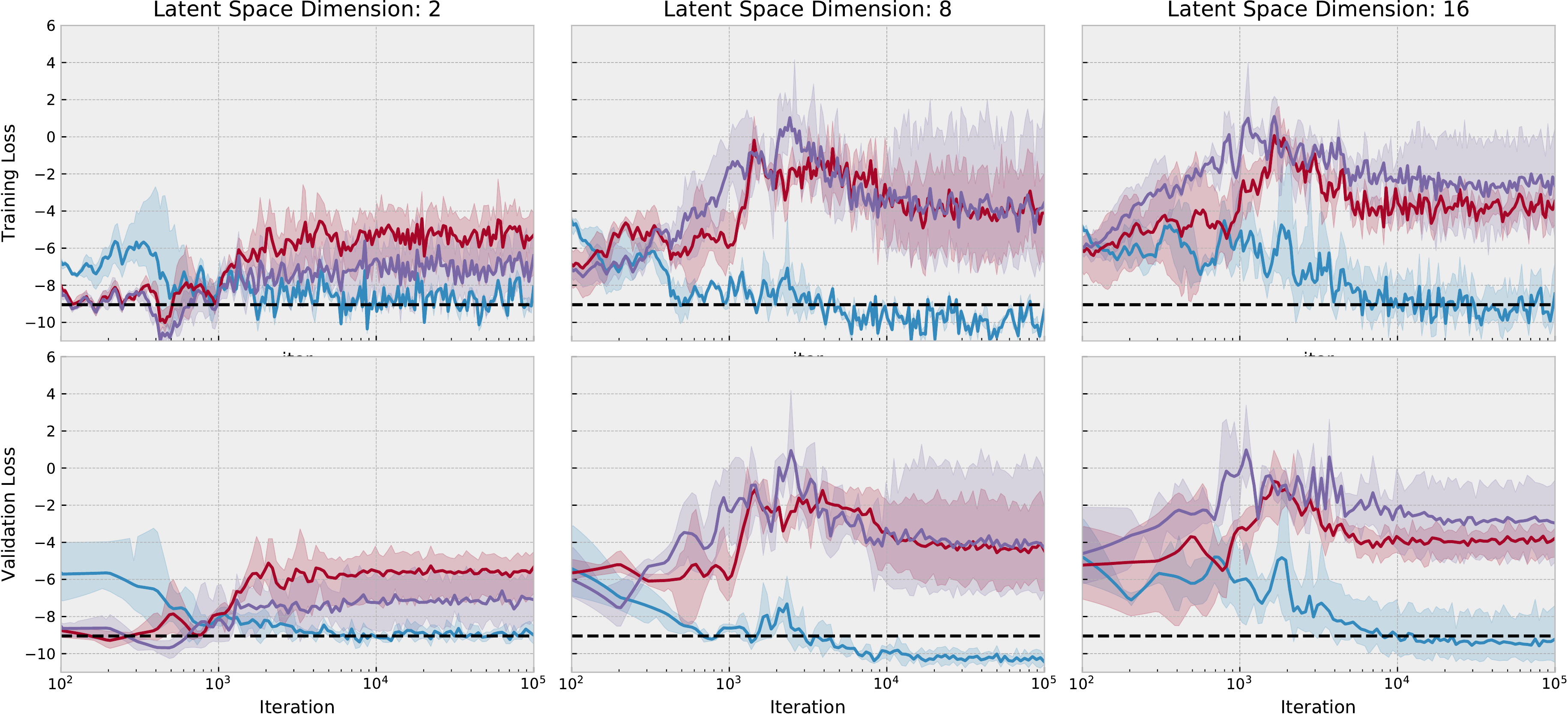}
  $\tau = $ (\cblock{83}{123}{164} 1.0 \cblock{187}{64}{60} 0.1 \cblock{159}{92}{149} 0.0)
  \caption{
    Training and validation loss convergence for the cartpole task.
    The dashed horizontal line shows the loss induced by an
    expert controller.
    Larger latent spaces seem harder to learn and as DCEM becomes
    less differentiable, the embedding is more difficult to learn.
    The shaded regions show the 95\% confidence interval around three trials.
  }
  \label{fig:cp:convergence}
\end{figure*}

For training, we sample initial starting states
of the cartpole and for validation we use a fixed set
of initial states.
\Cref{fig:cp:convergence} shows the convergence of
models as we vary the latent space dimension and
temperature parameter, and \cref{fig:cp:improve}
shows that DCEM is able to fully recover the
expert performance on the cartpole.
Because we are operating in the ground-truth dynamics
setting we measure the performance by comparing
the controller costs.
We use $\tau=0$ to indicate the case where we optimize
over the latent space with vanilla CEM and then
update the decoder with $\nabla_z C(\fdec_\theta(\hat z))$,
where the gradient doesn't go back into the optimization
process that produced $\hat z$.
This is non-convex min differentiation and is reasonable
when $\hat z$ is near-optimal, but otherwise is susceptible
to making the decoder difficult to search over.

These results show a few interesting points that come up in
this setting, which may be different in other settings.
Firstly that with a two-dimensional latent space, all
of the temperature values are able to find a reasonable latent
space at some point during training.
However after more updates, the lower-temperature experiments
start updating the decoder in ways that make it more difficult
to search over and start achieving worse performance
than the $\tau=1$ case.
For higher-dimensional latent spaces, the DCEM machinery
is necessary to keep the decoder searchable.
We notice that just a 16-dimensional latent space
for this task can be difficult for learning, one reason
this could be is from DCEM having too many degrees of freedom
in ways it can update the decoder to improve the performance
of the optimizer.

\section{More details: Cheetah and walker experiments}
\label{app:control-suite}

\begin{figure}[t]
\centering
\scalebox{0.95}{\tikzset{>=latex}
\begin{tikzpicture}[
deterministic_variable/.style={rectangle, draw=black, fill=white, thin, minimum width=20pt, minimum height=20pt},
invisible_variable/.style={rectangle, draw=white, fill=white, thin, minimum width=20pt, minimum height=20pt},
stochastic_variable/.style={circle, draw=black, fill=white,  minimum size=20pt},
observable_variable/.style={circle, draw=black, fill=gray!25,  minimum size=20pt},
]

\node[stochastic_variable] (z) {$z$};

\node[deterministic_variable, below left =60pt and 50pt of z.south] (h_1) {$h_1$};
\node[deterministic_variable, below left =60pt and 10pt of z.south] (h_2) {$h_2$};
\node[deterministic_variable, below right=60pt and 10pt of z.south] (h_3) {$h_3$};
\node[invisible_variable,     below right=60pt and 50pt of z.south] (h_4) {$\cdot\cdot\cdot$};

\node[stochastic_variable, above=20pt of h_1.north] (u_1) {$u_1$};
\node[stochastic_variable, above=20pt of h_2.north] (u_2) {$u_2$};
\node[stochastic_variable, above=20pt of h_3.north] (u_3) {$u_3$};
\node[invisible_variable,  above=20pt of h_4.north] (u_4) {};

\node[observable_variable, below=20pt of h_1.south] (x_1) {$x_1$};
\node[observable_variable, below=20pt of h_2.south] (x_2) {$x_2$};
\node[observable_variable, below=20pt of h_3.south] (x_3) {$x_3$};

\node[observable_variable, below=60pt of h_1.south] (r_1) {$r_1$};
\node[observable_variable, below=60pt of h_2.south] (r_2) {$r_2$};
\node[observable_variable, below=60pt of h_3.south] (r_3) {$r_3$};

\draw[->] (z.south) -- (u_1.north);
\draw[->] (z.south) -- (u_2.north);
\draw[->] (z.south) -- (u_3.north);
\draw[->] (z.south) -- (u_4.north);

\draw[->] (u_1.south) -- (h_2.north);
\draw[->] (u_2.south) -- (h_3.north);
\draw[->] (u_3.south) -- (h_4.north);

\draw[->] (h_1.east) -- (h_2.west);
\draw[->] (h_2.east) -- (h_3.west);
\draw[->] (h_3.east) -- (h_4.west);

\draw[->] (h_1.south) -- (x_1.north);
\draw[->] (h_2.south) -- (x_2.north);
\draw[->] (h_3.south) -- (x_3.north);

\path[->] (h_1.south east) edge [out=300, in=60] (r_1.north east);
\path[->] (h_2.south east) edge [out=300, in=60] (r_2.north east);
\path[->] (h_3.south east) edge [out=300, in=60] (r_3.north east);

\draw[->] (x_1.north east) -- (h_2.south west);
\draw[->] (x_2.north east) -- (h_3.south west);
\draw[->] (x_3.north east) -- (h_4.south west);

\draw[->] (x_1.south) -- (r_1.north);
\draw[->] (x_2.south) -- (r_2.north);
\draw[->] (x_3.south) -- (r_3.north);

\end{tikzpicture}}
\caption{Our RSSM with action sequence embeddings}
\label{fig:rssm}
\end{figure}
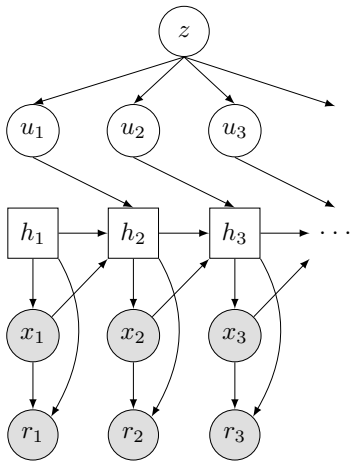

For the \verb!cheetah.run! and \verb!walker.walk! DeepMind control suite
experiments we start with a modified PlaNet \citep{hafner2018learning}
architecture without a pixel decoder.
We started with this over PETS \citep{chua2018deep}
to show that this RSSM is reasonable for proprioceptive-based
control and not just pixel-based control.
This model is graphically shown in \cref{fig:rssm} and has
1) a deterministic state model $h_t=f(h_{t-1}, x_{t-1}, u_{t-1})$,
2) a stochastic state model $x_t \sim p(x_t, h_t)$,
and 3) a reward model: $r_t\sim p(r_t | h_t, x_t)$.
In the proprioceptive setting, we posit that the deterministic
state model is useful for multi-step training even in
fully observable environments as it allows the model to
``push forward'' information about what is potentially
going to happen in the future.

For the modeling components, we follow the recommendations
in \citet{hafner2018learning} and use a GRU \citep{cho2014learning}
with 200 units as the deterministic path in the dynamics model and
implement all other functions as two fully-connected layers,
also with 200 units with ReLU activations.
Distributions over the state space are isotropic Gaussians
with predicted mean and standard deviation.
We train the model to optimize the variational bound on the
multi-step likelihood as presented in \citep{hafner2018learning}
on batches of size 50 with trajectory sequences of length 50.
We start with 5 seed episodes with random actions and in contrast
to \citet{hafner2018learning}, we have found that
interleaving the model updates with the environment steps
instead of separating the updates slightly improves the
performance, even in the pixel-based case, which we do not
report results on here.

For the optimizers we either use CEM over the full control space
or DCEM over the latent control space and use a horizon length
of 12 and 10 iterations here. For full CEM, we sample
1000 candidates in each iteration with 100 elite candidates.
For DCEM we use 100 candidates in each iteration with
10 elite candidates.

Our training procedure has the following three phases, which we
set up to isolate the DCEM additions.
We evaluate the models output from these training runs on 100
random episodes in \cref{fig:dmc:rew} in the main paper.
Now that these ideas
have been validated, promising directions of future work include
trying to combine them all into a single training run and
trying to reduce the sample complexity and number of timesteps
needed to obtain the final model.

\textbf{Phase 1: Model initialization.}
We start in both environments by launching a single training
run of \cref{alg:planet-proprio} to get initial system dynamics.
These models take slightly longer to converge
than in \citep{hafner2018learning}, likely due to how
often we update our models.
We note that at this point, it would be ideal to use the
policy loss to help fine-tune the components so that
policy induced by CEM on top of the models can be guided,
but this is not feasible to do by backpropagating through
all of the CEM samples due to memory, so we instead
next move on to initializing a differentiable controller
that is feasible to backprop through.

\textbf{Phase 2: Embedded DCEM initialization.}
Our goal in this phase is to obtain a differentiable controller
that is feasible to backprop through.

Our first failed attempt to achieve this was to use offline
training on the replay buffer, which would have been ideal
as it would require no additional transitions to be collected
from the environment.
We tried using \cref{alg:embed}, the same procedure we used
in the ground-truth cartpole setting, to generate an embedded
DCEM controller that achieves the same control cost on the replay
buffer as the full CEM controller.
However we found that when deploying this controller on the
system, it quickly stepped off of the data manifold and
failed to control it --- this seemed to be from the controller
finding holes in the model that causes the reward to
be over-predicted.

We then used an online data collection process identical to
the one we used for phase 1 to jointly learn the embedded
control space while updating the models so that the embedded
controller doesn't find bad regions in them.
We show where the DCEM updates fit into \cref{alg:planet-proprio}.
One alternative that we tried to updating the decoder to
optimize the control cost on the samples from the
replay buffer is that the decoder can also be immediately
updated after planning at every step.
This seemed nice since it didn't require any additional
DCEM solves, but we found that the decoder became too
biased during the episode as samples at consecutive
timesteps have nearly identical information.
For the hyper-parameters, we kept most of the DCEM hyper-parameters
fixed throughout this phase to 100 samples, 10 elites,
and a temperature $\tau=1$.
We ablated across 1) the number of DCEM iterations taken
to be $\{3, 5, 10\}$, 2) deleting the replay buffer
from phase 1 or not, and 3) re-initializing the
model or not from phase 1.

\textbf{Phase 3: Policy optimization into the controller.}
Finally now that we have a differentiable policy class induced
by this differentiable controller we can do policy learning to
fine-tune parts of it.
We initially chose Proximal Policy Optimization (PPO) \citep{schulman2017proximal}
for this phase because we thought that it would be able to fine-tune
the policy in a few iterations without requiring a good
estimate of the value function, but this phase also ended up
consuming many timesteps from the environment.
Crucially in this phase, we do \textbf{not} do likelihood
fitting at all, as our goal is to show that PPO can be used
as another useful signal to update the parts of a controller ---
we did this to isolate the improvement from PPO but in practice
we envision more unified algorithms that use both
signals at the same time.
Using the standard PPO hyper-parameters, we collect 10
episodes for each PPO training step and ablate across
1) the number of passes to make through these episodes $\{1, 2, 4\}$,
2) every combination of the reward, transition, and decoder
being fine-tuned or frozen,
3) using a fixed variance of 0.1 around the output of
the controller or learning this,
4) the learning rate of the fine-tuned model-based portions
$\{10^{-4}, 10^{-5}\}$.

\begin{algorithm*}[t]
  \caption{PlaNet \citep{hafner2018learning} variant that we use for proprioceptive control
    with optional DCEM embedding}
\label{alg:planet-proprio}
\begin{algorithmic}
  \LeftComment \textbf{Models:}
  a deterministic state model, a stochastic state model,
  a reward model, and (if using DCEM) an action sequence decoder.
  \LeftComment Initialize dataset $\gD$ with $S$ random seed episodes.
  \LeftComment Initialize the transition model's deterministic hidden
  state $h_0$ and initialize the environment, obtaining the
  initial state estimate $x_0$.
  \LeftComment $\text{CEM-Solve}$ can use DCEM or full CEM
  \For {$t = 1, \ldots, T$}
  \State $u_t \leftarrow \text{CEM-solve}(h_{t-1}, x_{t-1})$
  \State Add exploration noise $\epsilon\sim p(\epsilon)$ to the action $u_t$.
  \State $\{r_t, x_{t+1}, d_t\} \leftarrow \text{env.step}(u_t)$ \Comment Properly restarting if necessary
  \State Add $[r_t, x_t, u_t, d_t]$ to $\gD$
  \State $h_t=\text{update-hidden}(h_{t-1}, x_t, u_t, d_t)$
  \If {$t \equiv 0 \pmod{\text{update-interval}}$}
    \State Sample trajectories $\tau = [r_\tau, x_\tau, u_\tau, d_\tau]_{\tau=1}^H \sim \gD$ from the dataset.
    \State Obtain the hidden states of the $\{h_\tau,\hat x_\tau\}$ from the model.
    \State Compute the multi-step likelihood bound $\Ls(\tau, h_\tau, \hat x_\tau)$ \Comment Eq.~6 of \citet{hafner2018learning}
    \State $\theta \leftarrow \text{grad-update}(\nabla_\theta \Ls_\theta(\tau, h_\tau, \hat x_\tau))$
    \Comment Optimize the likelihood bound
    \If {using DCEM}
      \State $\hat z_\tau = \argmin_{z\in \gZ} C_\theta(z; h_\tau, \hat x_\tau)$
      \Comment Solve the embedded control problem in \cref{eq:emb_ctrl}
      \State $\theta \leftarrow \text{grad-update}(\nabla_\theta \sum_\tau C_\theta(\hat z_\tau))$
    \Comment Update the decoder
    \EndIf
  \EndIf
  \EndFor
\end{algorithmic}
\end{algorithm*}

\begin{figure*}[ht]
  \centering
  \includegraphics[width=0.8\textwidth]{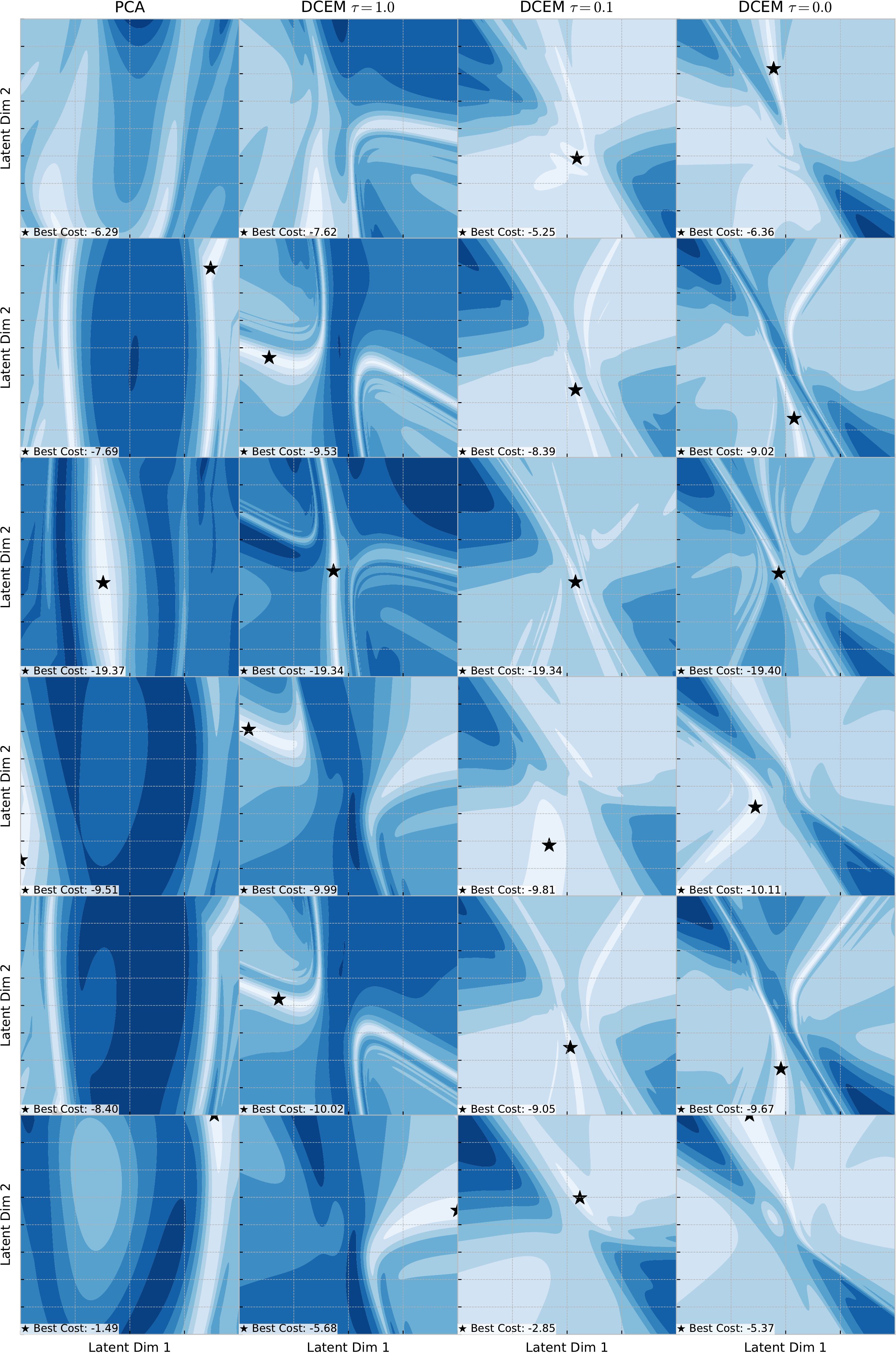}
  \caption{Learned DCEM reward surfaces for the cartpole task.
    Each row shows a different initial state of the system.
    We can see that as the temperature decreases, the latent
    representation can still capture near-optimal values,
    but they are in much narrower regions of the latent space
    than when $\tau=1$.
  }
  \label{fig:cp:surfaces}
\end{figure*}

\end{document}